\documentclass{article}
\usepackage{spconf,amsmath,graphicx,amsfonts}
\usepackage{subfigure}
\usepackage{algorithm}
\usepackage{algorithmicx}
\usepackage{algpseudocode}
\usepackage{booktabs}
\usepackage{multirow}
\usepackage{url}

\newtheorem{myDef}{Definition}
\title{Efficient Divide-And-Conquer Classification Based on Feature-Space Decomposition}
\name{Qi Guo$^1$, Bo-Wei Chen$^{2*}$, Feng Jiang$^3$, Xiangyang Ji$^1$ and Sun-Yuan Kung$^2$}
\address{$^1$Tsinghua University, Beijing 100084, China,\\
              $^2$Princeton University, Princeton, NJ 08544, USA (email: dennisbwc@gmail.com)\\
              $^3$Harbin Institute of Technology, Harbin 150001, China\\}

\begin{document}
%\ninept
%
\maketitle
\begin{abstract}
This study presents a divide-and-conquer (DC) approach based on feature space decomposition for classification. When large-scale datasets are present, typical approaches usually employed truncated kernel methods on the feature space or DC approaches on the sample space. % NOT SURE ABOUT THE EXPRESSION
 However, this did not guarantee separability between classes, owing to overfitting. To overcome such problems, this work proposes a novel DC approach on feature spaces consisting of three steps. Firstly, we divide the feature space into several subspaces using the decomposition method proposed in this paper. Subsequently, these feature subspaces are sent into individual local classifiers for training. Finally, the outcomes of local classifiers are fused together to generate the final classification results. Experiments on large-scale datasets are carried out for performance evaluation. The results show that the error rates of the proposed DC method decreased comparing with the state-of-the-art fast SVM solvers, e.g., reducing error rates by 10.53\% and 7.53\% on RCV1 and covtype datasets respectively.
\end{abstract}
\section{Introduction}
\label{sec:intro}
Typical kernel-based classification, such as Support Vector Machines (SVMs) \cite{vapnik1998statistical} and Kernel Ridge Regression (KRR) \cite{cristianini2000introduction}, usually employs Radial Basis Functions (RBFs) as the kernel, for RBFs can effectively delineate the distribution of the data by using mixtures of Gaussian models. Furthermore, RBFs can map the input features into the intrinsic space \cite{kung2014kernel} that is spanned by infinite-dimensional vectors. This correspondingly increases the opportunity of creating a discriminant hyperplane in the empirical space \cite{kung2014kernel}, subsequently enhancing discriminability. However, when input dimensions are sufficiently large, calculation of a kernel matrix becomes a burden.  Moreover, RBFs may lead to overfitting  due to infinite dimensions. To deal with such problems, rather than using conventional RBFs, Wu \emph{et al}. \cite{kung2012efficient} proposed using Truncated Radical Basis Functions (TRBFs) to avoid generating infinite dimensions in the intrinsic space. Furthermore, they also devised an intrinsic data matrix, which was derived from a finite-decomposable kernel, to replace calculation of kernel matrices in the empirical space. Therefore, the time complexity was saved from original $O(N^3)$ to $min(N^3,J^2N+J^3)$ for KRR, where $N$ is the number of instances, and $J$ is the number of feature dimension expanded by TRBFs. Moreover, avoiding direct calculation of kernel matrices effectively resolved the need for matrix expansion.

The success of TRBF-based method relies on dimensional reduction in the intrinsic space and the conversion from empirical space to intrinsic space. Although computational load is relieved without losing too much accuracy, however, that method \cite{kung2012efficient} did not improve discriminability and separability between features. Furthermore, the algorithmic architecture of that method did not support distributed processing, especially when mainstream toolboxes like Apach Hadoop (hadoop.apache.org) and Spark (spark.apache.org) adopt divide-and-conquer strategy in their implementation. Proposing a new architecture that supports divide-and-conquer computation correspondingly becomes necessary.

In response to such a demand, several divide-and-conquer classifiers \cite{zhang2013divide},\cite{hsieh2013divide} based on kernel tricks have been developed so far. Zhang \emph{et al}. used divide-and-conquer KRR \cite{zhang2013divide} to support computation of large-scale data. Firstly, their method randomly partitioned a dataset into subsets of equal size. Local solutions were subsequently computed by using KRR based on each subset. By averaging the local solutions, a global predictor was therefore obtained. Instead of using randomized data selection as Zhang \emph{et al}. did, Hsieh \emph{et al}. \cite{hsieh2013divide} focused on systematic data division before applying divide-and-conquer classifiers to the data. In their approach, kernel K-means clustering was performed to select the representatives of the entire input data. Next, the members of a subset were selected based on one representative. Their experimental result showed a favorable accuracy when systematic data division was used.

Although the above-mentioned approaches realized divide-and-conquer concept in their algorithms, overfitting of kernel space was not fully addressed and resolved. To deal with the aforementioned problems, this study proposes

%
%
% INDEPENDENT SUBSPACE????
1) A novel approach for feature-space decomposition, where the original feature space is converted to subspaces. Besides, the bases of each subspace are reranked according to their importance.

2) A divide-and-conquer structure that allows independent local classifiers to create discriminant hyperplanes based subspaces rather than the entire empirical space. This lowers computational complexity while avoiding overfitting problems.

The rest of this paper is organized as follows. Section \ref{sec:sys} introduces the overview of the proposed method. Section \ref{sec:method} then describes details of the proposed feature-space decomposition and fusion method. Next, Section \ref{sec:exp} summarizes the performance of the proposed method and the analysis results. Conclusions are finally drawn in Section \ref{sec:conclude}.
%% PAPER STRUCTURE

\vspace{-0.5em}
\section{System Overview}
\label{sec:sys}
Given an $M{\times}N$ data matrix $X$ with $N$ instances and $M$ features and a $1{\times}N$ label vector $y$, denote the feature space as $\Omega$, and $X$ are the projection of the $N$ instances on $\Omega$. We first define the feature-space decomposition method $D = \{T, I\}$, where $T$ is a feature-space transform function, and $I$ is a set of feature index groups.
% of features that are put together to span the $i$th feature subspace and $m$ is the number of subspaces.

The decomposition method $D$ contains five sub-methods which are discussed in Section \ref{sec:dec}, namely, Random Decomposition (RD), Principle Component Analysis (PCA), Discriminant Component Analysis (DCA), Block Cholesky Decomposition (BCD) and Approximate Block Decomposition (ABD). Furthermore, each have an $M\times M$ sub-transform matrix, denoted as $W^{RD}$, $W^{PCA}$, $W^{DCA}$, $W^{BCD}$ and $W^{ABD}$. Also, each contains a subset of feature index groups, e.g., $I^{RD}=\{I^{RD}_i|I^{RD}_i\subset \{1,2,\cdots,M\}, i=1,2,\cdots,h^{RD}\}$, where $h^{RD}$ is the number of feature subspaces decomposed by RD sub-method, respectively. As for $T$, we have
\begin{equation}
\label{eqn:transX}
\Omega^{\ast}=T(\Omega), X^{\ast}=T(X)=WX=
\left(
\footnotesize
\begin{array} {c}
W^{RD} \\
W^{PCA} \\
W^{DCA} \\
W^{BCD} \\
W^{ABD} \\
\end{array}
\normalsize
\right)X
\end{equation}
where $W$ and $\Omega^{\ast}$ are respectively the transform matrix and the new feature space. As for $I$, we have $I=\{I^{RD}$, $I^{PCA}$, $\cdots$, $I^{ABD}\}$ , and the total number of subspaces is $h=h^{RD}+h^{PCA}+\cdots+h^{ABD}$. Not all the sub-methods need to be used in real practice. If some are not applied, the corresponding $W_{Method}$ and $I_{Method}$ can just be empty.

The original feature space $\Omega$ is first transformed to $\Omega^{\ast}$ by $T$ and then decomposed into subspaces $\Omega^{\ast}_1,\Omega^{\ast}_2,...,\Omega^{\ast}_h$ by $I$; all the instances are first projected $X^{\ast}$ and subsequently decomposed into $X^{\ast}_1,X^{\ast}_2 ,..., X^{\ast}_h$.Then, a local classifier $f_i$($i=1,2,\cdots,h$), e.g., SVMs, KRRs, etc. is trained using data matrix $X^{\ast}_i$. Let row vectors $R_i = f_i(X^{\ast}_i)$ denote the results of $f_i$ on $X^{\ast}_i$ and
$R =
\footnotesize
\left(
\begin{array} {c}
R_1 \\
R_2 \\
\vdots \\
R_h\\
\end{array}
\right)
\normalsize
$
. The elements of $R_i$ can be discrete labels or continuous prediction values. The system generates the output based on $R$ using fusion methods which are discussed in Section \ref{sec:fusion}.

\vspace{-1em}
\section{Proposed Division and Fusion Methods}
\label{sec:method}
\vspace{-0.5em}
\subsection{Feature-Space Decomposition}
{\label{sec:dec}
Section 2 shows that the merit of the proposed method is, it can perform classification within the subspaces and ignores the dependance among subspaces. Theoretically, decomposition method should be able to reduce the dependance as much as possible between any two feature subspaces while remaining dependance within the subspaces. This is the reason for conducting transformation on the feature space before division.

Among all the sub-methods in this study, the simplest idea is RD which directly decompose the feature space based on $I$. Its $W_{RD}$ is an $M \times M$ identity matrix.

As for PCA, we conduct PCA on the data matrix $X$ and split up the features according to $I$.  Since PCA diagonizes the feature covariance matrix $\overline{S}$, this method eliminates the relevance of different features among and within subspaces. If the data obey Gaussian distribution, the PCA also eliminate the dependance of features among and within feature subspaces.

DCA also conducts orthogonal transformation like PCA,
while its discriminant matrix is $[S_w+\rho I]^{-1}\overline{S}$ , where $S_w$ is the within-class scatter matrix, and $\rho$ is the ridge parameter \cite{kung2014kernel}. We have
\begin{equation}
\label{eqn:dca}
S_w =\Sigma_{l=1}^{L}\Sigma_{j=1}^{N_l}[x^{(l)}_j-\mu^{(l)}][x^{(l)}_j-\mu^{(l)}]^T
\end{equation}
where $l$ is the number of classes, $N_l$ represents the number of samples in class $l$, and $\mu^{(l)}$ specifies the average point of class. We conduct generalized eigenvector decomposition \cite{moler1973algorithm} to obtain the eigenvectors $\nu_1,\nu_2,...,\nu_M$and eigenvalue matrix $\lambda_1,\lambda_2,...,\lambda_M$, such that
\begin{equation}
\label{eqn:geig}
\overline{S}\nu_k = \lambda_k[S_w+\rho I]\nu_k,k=1,2,...,M
\end{equation}
and the transform matrix is defined as $W_{DCA} = $ $[\nu_1$, $\nu_2$, $\cdots$, $\nu_M]$. Computing $S$ and $S_w$ both enjoys $O(M^2N)$ complexity. As $[S_w+\rho I]$ can hardly be singular, the complexity of generalized eigenvalue decomposition equals that of $\lambda_k[S_w+\rho I]^{-1}\overline{S}\nu_k = \lambda_k\nu_k,k=1,2,...,M$, which is of $O(M^3)$ time complexity. Therefore, the total complexity of DCA is $O(2M^2N+M^3)$.

BCD exploits a blocked Doolittle Algorithm, which is a form of Gaussian transformation rather than the orthogonal transformation, to eliminate the relevance among subspaces while remaining relevance within subspaces. For a symmetrical block matrix $A$
, we eliminate the first row and column of blocks, as shown in Equation \ref{eqn:doolittle}.
\begin{equation}\label{eqn:doolittle}
\footnotesize
\left(
\begin{array} {cccc}
A'_{11} & O_{12} & \cdots & O_{k1}^T \\
O_{21}& A'_{22} & \cdots & {A'_{k2}}^T \\
\vdots & \vdots & \ddots & \vdots \\
O_{k1} & A'_{k2} & \cdots & A'_{kk} \\
\end{array}
\right)
=B^1
\left(
\begin{array} {ccc}
A_{11} & \cdots & A_{k1}^T \\
\vdots & \ddots & \vdots \\
A_{k1} & \cdots & A_{kk} \\
\end{array}
\right)
({B^1})^T
\normalsize
\end{equation}
where $B_1=
\footnotesize
\left(
\begin{array} {cccc}
I_{11} & O_{12} & \cdots & O_{1k} \\
-A_{21}A_{11}^{-1} & I_{22} & \cdots & O_{2k} \\
\vdots & \vdots & \ddots & \vdots \\
-A_{k1}A_{11}^{-1} & O_{k2} & \cdots & I_{kk}\\
\end{array}
\right)
\normalsize
$, and the division of blocks remains the same. The subscript of $B$ indicates the row and column it eliminates. Iteratively, we subsequently generate $B_2$,$\cdots$,$B_k$ to sequentially eliminate the rest rows and columns of blocks. The main goal of BCD is sequentially block-diagonizing the discriminant matrix based on the blocked Doolittle Algorithm. As shown in Algorithm \ref{alg:BCD}, $X$ is firstly rearranged to generate $\tilde{X}$ according to $I^{BCD}$, in which MatrixSplit$(X,I^{BCD})$ splits $X$ into  $\tilde{X_1}$, $\tilde{X_2}$, $\cdots$ , $\tilde{X}_{h^{BCD}}$ according to $I^{BCD}$. The discriminant matrix of BCD is $\overline{S}$. Function BlockedDoolittle$(\tilde{X},I)$ generates $B_i$$(i=1,2,\cdots,h^{BCD})$ based on the idea of Equation \ref{eqn:doolittle} to eliminate the $i$th row and column of $\overline{S}$. The BCD transform matrix is $W_{BCD}=B_{h^{BCD}}B_{h^{BCD}-1}\cdots B_{1}$. Comparing to BCD, PCA needs to do an $M\times M$ matrix inversion, whose complexity is $O(M^3)$ on non-sparse matrix, whereas BCD only uses an $\frac{M}{m}\times\frac{M}{m}$ matrix for $\frac{m(m+1)}{2}$ times if divided equally, which only costs $\frac{1}{m}$ of the time of PCA.
\begin{algorithm}\caption{$[X^{\ast}, W_{BCD}] = \mathrm{BCD}(X,I)$}
\label{alg:BCD}
\begin{algorithmic}
%\Require $X$: the original $M{\times}N$ data matrix, $I$: the feature division groups
%\Ensure $\delta$:the classification result, where 1 represents open eye,0 represents closed eye.\\
\footnotesize
\State $\{\tilde{X_1},\tilde{X_2},...,\tilde{X_n}\}=$MatrixSplit$(X,I)$
\State $\tilde{X}=[\tilde{X_1},\tilde{X_2},...,\tilde{X_m}]$
\State $\overline{S}=\tilde{X}\tilde{X}^T=\left(
\begin{array} {ccc}
\overline{S_{11}} & \cdots & \overline{S_{1m}} \\
\vdots & \ddots & \vdots \\
\overline{S_{m1}} & \cdots & \overline{S_{mm}} \\
\end{array}
\right) $
, where $\overline{S_{ij}}=\tilde{X_i}\tilde{X_j}^T$
\For{$i = 1$ to $h^{BCD}$}
\State $B_{i}=$BlockedDoolittle$(\overline{S},i)$
\State $\overline{S}=B_{i}\overline{S}(B_{i})^T $
\EndFor
\State $W^{BCD} = B_{h^{BCD}}B_{h^{BCD}-1}\cdots B_{1}$
\State $X^{\ast}=W^{BCD}\tilde{X}$
\normalsize
\end{algorithmic}
\end{algorithm}
% PROVE WHY THE FIRST BLOCK CONTAINS THE MOST INFORMATION

Besides the aforementioned orthogonal transformation of PCA and DCA, as well as the Gaussian transformation of BCD, we also propose an approximate orthogonal transformation on which the ABD method is based. First we define a new operator $\otimes$ as Definition \ref{def:rtime}.
\begin{myDef}
\label{def:rtime}
For two blocked matrix $A=\{A_{ij}\}$ and $B=\{B_{ij}\}$ with the same size and division of blocks, define
operator $\otimes$, s.t.
\begin{equation}\label{eqn:otime}
  \footnotesize
A\otimes B=
\left(
\begin{array} {ccc}
x_{11} & \cdots & x_{1m} \\
\vdots & \ddots & \vdots \\
x_{m1} & \cdots & x_{mm} \\
\end{array}
\right)
\normalsize
\end{equation}
where $x_{ij}$ equals the sum of all the elements of $A_{ij}$ element-wise multiply $B_{ij}$.
\end{myDef}

We rewrite X as
$
\footnotesize
\left(
\begin{array} {ccc}
X_{11} & \cdots & X_{1N} \\
\vdots & \ddots & \vdots \\
X_{h^{ABD}1} & \cdots & X_{h^{ABD}N} \\
\end{array}
\right)
\normalsize
$
where we divide each instance into $m$ vectors according to $I$. The discriminant matrix is $X \otimes X$ using this division. By conducting eigenvector decomposition on the discriminant matrix, we have
\begin{equation}\label{eqn:ABD}
X \otimes X=
V^T
\Lambda
V
\end{equation}
where $V=\{v_{ij}\}$, and each column of $V$ is an eigenvector. The transform matrix is
\begin{equation}
\label{eqn:ABDW}
W_{ABD}=
\footnotesize
\left(
\begin{array} {ccc}
v_{11}I_{11} & \cdots & v_{1N}I_{1N} \\
\vdots & \ddots & \vdots \\
v_{m1}I_{m1} & \cdots & v_{mN}I_{mN} \\
\end{array}
\right).
\normalsize
\end{equation}
If there are approximately equal number of features in each subset, computing $X \otimes X$ yields $O(MmN)$ complexity and the eigenvalue decomposition costs $O(m^3)$. Therefore, the total complexity of ABD is $O(MmN+m^3)$.

% Need to be re-written
Table \ref{table:comp} shows the time complexity and details of the aforementioned sub-methods. By combining the five methods together, $D$ includes both supervision (i.e., DCA) and unsupervision (i.e., RD,PCA,BCD and ABD) in transformation as well as four transformation methods.
%%I am not sure how to express
\begin{table}
\footnotesize
    \caption{Time complexity of different transformation methods.}
    \vspace{-1.5em}
    \label{table:comp}
    \begin{center}
      \begin{tabular}{ l | c | c}
        \hline
         & Complexity & Detail \\ \hline
        RD & $O(N)$ & Unsupervised, identity transform \\ \hline
        PCA & $ O(M^2N+M^3)$ & Unsupervised, orthogonal transform (OT) \\ \hline
        DCA & $ O(2M^2N+M^3)$ & Supervised, OT \\ \hline
        BCD & $ O(M^2N+M^3/m)$ & Unsupervised, Gaussian transform \\ \hline
        ABD & $O(MmN+m^3)$ & Unsupervised, approximate OT \\ \hline
      \end{tabular}
    \end{center}
\normalsize
\vspace{-2em}
\end{table}
}

\subsection{Feature Subspace Fusion}
{\label{sec:fusion}
After obtaining the classification result matrix $R$ from local classifier, we weight the outcome of each subspace by training a global classifier $f_{n+1}$ by using $R$ as a data matrix and $y$ as labels. The output of $f_{n+1}$ is the final prediction result. Observations show that $m<50<<N$ and TRBFKRR\cite{kung2012efficient} generates favorable results for $f_{n+1}$. As the training complexity of TRBFKRR is $min(N^3,J^2N+J^3)$, where
$J=
\left(
\begin{array} {c}
m+p \\
p \\
\end{array}
\right)
$
, and $p$ is TRBF order. It is efficient to train on data matrix with a large number of instances and a small number of features like $R$.}

\begin{table*}
    \caption{Decomposition setting. The $N_S$ and $N_F$ stand for number of subspaces and number of features in one subspace.}
    \vspace{0.5em}
    \footnotesize
    \label{table:decomp}
    \begin{center}
      \begin{tabular*}{15.2cm}{ l| p{3.2cm} | p{1.1cm} | p{1.1cm} | p{1.1cm} | p{1.1cm} | p{1.1cm} | p{1.1cm} }
        \hline
        Dataset & Proposed Method & Settings & RD & PCA & DCA & BCD & ABD \\ \hline
        \multirow{2}{2cm}{news20} & \multirow{2}{3.2cm}{DC-Liblinear-TRBF2KRR} & $N_S$ & $2$ & $0$ & $0$ & $0$ & $10$ \\ \cline{3-8}
        & & $N_F$ & $677596$ & $0$ & $0$ & $0$ & $135519$ \\ \hline
        \multirow{2}{2cm}{RCV1} & \multirow{2}{3.2cm}{DC-Liblinear-TRBF3RR} & $N_S$ & $4$ & $0$ & $0$ & $0$ & $4$ \\ \cline{3-8}
        & & $N_F$ & $23618$ & $0$ & $0$ & $0$ & $23618$ \\ \hline
        \multirow{2}{2cm}{covtype} & \multirow{2}{3.2cm}{DC-DCSVM-TRBF2KRR} & $N_S$ & $4$ & $4$ & $4$ & $4$ & $4$ \\ \cline{3-8}
        & & $N_F$ & $40$ & $40$ & $40$ & $27$ & $27$ \\ \hline
        \multirow{2}{2cm}{census} & \multirow{2}{3.2cm}{DC-DCSVM-LibLinear} & $N_S$ & $2$ & $2$ & $0$ & $0$ & $0$ \\ \cline{3-8}
        & & $N_F$ & $300$ & $300$ & $0$ & $0$ & $0$ \\ \hline
      \end{tabular*}
    \end{center}
    \normalsize
    \vspace{-3em}
\end{table*}

\section{Experimental Result}
\label{sec:exp}
In this section, we use LibLinear \cite{Fan:2008:LLL:1390681.1442794} and DCSVM \cite{hsieh2013divide} as local classifiers $f_1$,$f_1$,$\cdots$,$f_h$ in our system respectively and tested the results on large scale datasets (i.e., either $M$ or $N$ is larger than $10^4$). Our methods are notated as ``DC-classifier1-classifier2", where ``classifier1" indicates the classifier used for $f_1$,$f_1$,$\cdots$,$f_m$, and ``classifier2" is for $f_{n+1}$.
All the experiments are conducted on an Intel Core i7 2.1GHz CPU and 8G RAM machine. The datasets tested in this paper are shown in Table \ref{tab:datainfo} and can be downloaded from \url{http://www.csie.ntu.edu.tw/~cjlin/libsvmtools/datasets/} or UCI Machine Learning Repository.

\begin{table}
\footnotesize
    \caption{Dataset statistics. ``\#" represents ``number of". A random 0.9/0.1 split is applied to all news20 dataset. A random 0.8/0.2 split is applied to covtype and census dataset.}
    \vspace{-1em}
    \label{tab:datainfo}
    \begin{center}
      \begin{tabular}{ c | r | r | r }

        \hline
        Dataset & \# Training Instances & \# Testing Instances & \# Features \\ \hline
        news20 & $17,997$ & $1,999$ & $1,355,191$ \\
        RCV1 & $20,242$ & $677,399$ & $47,236$ \\
        covtype & $464,810$ & $116,202$ & $54$\\
        census & $159,619$ & $39,904$ & $409$\\\hline

      \end{tabular}
    \end{center}
\normalsize
\vspace{-2.5em}
\end{table}

\textbf{Feature-Space Decomposition Setting:} Table \ref{table:decomp} shows the decomposition setting in our experiments. For data matrices with high feature dimensions, e.g., news20, RCV1, we just use RD and ABD with relatively low computational complexity. For data matrices with low feature dimensions, all the transformation methods can be combined together to achieve a lower error rate. We use LibLinear as the global classifier when dealing with census dataset, as its proportion of positive and negative instances are 0.06/0.94, which can cause bias when TRBFKRR is applied.

\textbf{Linear Classification:} Linear classification is conducted on news20 and RCV1 datasets. LibLinear is exploited as local classifiers $f_1$,$f_2$,$\cdots$,$f_m$, and TRBFKRR is used as global classifier $f_{n+1}$ in our system. We compare our results with four common fast linear SVM solvers, namely, Liblinear \cite{Fan:2008:LLL:1390681.1442794}, SVMlight \cite{Joachims/99a}, BSVM \cite{6790353} and L2-SVM-MFN \cite{Sindhwani:2006:LSS:1148170.1148253}. As shown in Table \ref{tab:linclass}, our methods either have advantages on training times or error rates.
\begin{table}
\footnotesize
    \caption{Comparison of linear classification on real world datasets.}
    \vspace{-1.5em}
    \label{tab:linclass}
    \begin{center}
      \begin{tabular}{ c | c | c | c | c}
        \hline
         & \multicolumn{2}{c}{news20} & \multicolumn{2}{|c}{RCV1} \\ \cline{2-5}
         & Time (s) & Error Rate (\%) & Time (s) & Error Rate (\%) \\ \hline
        Proposed & $32$  & $2.87$ & $3.1$ & \boldmath$3.06$\unboldmath  \\ \hline
        LibLinear & \boldmath$2$\unboldmath & $3.26$ & \boldmath$0.3$\unboldmath & $3.84$ \\ \hline
        SVMlight & $467$\footnotemark[1] & $2.74$\footnotemark[1] & $15.5$ & $3.42$ \\ \hline
        BSVM & $437$\footnotemark[1] & \boldmath$2.73$\unboldmath\footnotemark[1] & $13.2$ & $3.68$ \\ \hline
        L2-SVM-MFN & $98$\footnotemark[1] & $2.86$\footnotemark[1] & $0.5$ & $3.53$ \\ \hline
      \end{tabular}
    \end{center}
\normalsize
\vspace{-2.5em}
\end{table}
\footnotetext[1]{Results are cited from Keerthi \emph{et al.} \cite{keerthi2005modified}}.
\footnotetext[2]{Results are cited from Hsieh \emph{et al.} \cite{hsieh2013divide}}

\textbf{Non-Linear Classification:} DCSVM is set as local classifiers for nonlinear classification. Interestingly, in DC-DCSVM-TRBFKRR, divide-and-conquer process is conducted on both instance dimension and feature dimension in the method. We evaluate it on covtype dataset and compare with the results of the other SVM methods by Hsieh \emph{et al.} \cite{hsieh2013divide}, as is shown in Table \ref{tab:nonlinclass}. The proposed method achieve the lowest error rate with relatively low time complexity in both covtype and census datasets.

\begin{table}
\footnotesize
    \caption{Comparison of nonlinear classification on real world datasets.}
    \vspace{-1em}
    \label{tab:nonlinclass}
    \begin{center}
      \begin{tabular}{ c | c | c | c | c}
        \hline
         & \multicolumn{2}{c}{covtype} & \multicolumn{2}{|c}{census} \\ \cline{2-5}
         & \multicolumn{2}{c|}{$c=32$,$\gamma=32$} &  \multicolumn{2}{c}{$c=512$,$\gamma=2^{-9}$}\\ \cline{2-5}
         & Time (s) & Error Rate (\%) & Time (s) & Error Rate (\%) \\ \hline
        Proposed & $7537$  & \boldmath$3.56$\unboldmath & $1459$ & \boldmath$5.0$\unboldmath  \\ \hline
        DCSVM(early)\footnotemark[2] & \boldmath$672$\unboldmath & $3.88$ & \boldmath$261$\unboldmath & $5.1$ \\ \hline
        DCSVM\footnotemark[2] & $11414$ & $3.85$ & $1051$ & $5.8$ \\ \hline
        LibSVM\footnotemark[2] & $83631$ & $3.85$ & $2920$ & $5.8$ \\ \hline
        LaSVM\footnotemark[2] & $102603$ & $5.61$ & $3514$ & $6.8$ \\ \hline
        CascadeSVM\footnotemark[2] & $5600$ & $10.49$ & $849$ & $7.0$ \\ \hline
        LLSVM\footnotemark[2] & $4451$ & $15.79$ & $1212$ & $7.2$ \\ \hline
        FastFood\footnotemark[2] & $8550$ & $19.9$ & $851$ & $8.4$ \\ \hline
        SpSVM\footnotemark[2] & $15113$ & $16.63$ & $3121$ & $9.6$ \\ \hline
        LTPU\footnotemark[2] & $11532$ & $16.75$ & $1695$ & $8.0$ \\ \hline
      \end{tabular}
    \end{center}
\normalsize
\vspace{-2.5em}
\end{table}

Moreover, comparing to directly training a TRBFKRR classifier using the whole data matrix, DC-TRBFKRR-TRBFKRR greatly reduces the training complexity from $min(N^3,J^2N+J^3)$ to $min(N^3,\frac{J^2N}{m}+\frac{J^3}{m^2})$, which enables TRBFKRR training on data matrix with large $N$ and $M$.

\section{Conclusion}
\label{sec:conclude}
This paper presents a feature-space decomposition classification method including five sub-methods. The experimental results show that our divide-and-conquer classification scheme can reduce error rates (e.g., reduce error rates by 10.53\% and 7.53\% in covtype and RCV1 datasets), comparing to training directly using the whole datasets, and outperform state-of-the-art fast SVM solver by reducing overfitting problem. The future work will focus on providing theoretical analysis for feature-space decomposition and its effects on divide-and-conquer classification.

\vfill\pagebreak

\bibliographystyle{IEEEbib}
\bibliography{strings,refs}
\end{document}